\documentclass{article}

\usepackage{times}
\usepackage{graphicx} 
\usepackage{subfigure} 

\usepackage{natbib}

\usepackage{multirow}
\usepackage{algorithm}
\usepackage{algorithmic}

\usepackage{hyperref}

\usepackage[accepted]{icml2017}

\usepackage{algorithm}

\usepackage{latexsym}
\usepackage[utf8]{inputenc} 
\usepackage[T1]{fontenc}   
\usepackage{hyperref}       
\usepackage{url}            
\usepackage{booktabs}       
\usepackage{amsfonts}       
\usepackage{nicefrac}       
\usepackage{microtype}      
\usepackage[centertags]{amsmath}
\usepackage{amssymb}
\usepackage{graphicx}
\usepackage{tabularx}
\usepackage{subfigure}
\usepackage{paralist}
\usepackage[inline,shortlabels]{enumitem}

\newcommand{\DD}{\mathcal{D}}

\newcommand{\cev}[1]{\reflectbox{\ensuremath{\vec{\reflectbox{\ensuremath{#1}}}}}}

\DeclareMathOperator*{\expect}{\mathbb{E}}

\DeclareMathOperator*{\maximize}{maximize}

\icmltitlerunning{Learning Algorithms for Active Learning}

\begin{document} 

\twocolumn[
\icmltitle{Learning Algorithms for Active Learning}



\icmlsetsymbol{equal}{*}

\begin{icmlauthorlist}
\icmlauthor{Philip Bachman}{equal,maluuba}
\icmlauthor{Alessandro Sordoni}{equal,maluuba}
\icmlauthor{Adam Trischler}{maluuba}
\end{icmlauthorlist}

\icmlaffiliation{maluuba}{Microsoft Maluuba, Montreal, Canada}

\icmlcorrespondingauthor{P.~ Bachman}{phbachma@microsoft.com}
\icmlcorrespondingauthor{A.~ Sordoni}{alsordon@microsoft.com}

\icmlkeywords{boring formatting information, machine learning, ICML}

\vskip 0.3in
]



\printAffiliationsAndNotice{\icmlEqualContribution} 

\begin{abstract}
We introduce a model that learns active learning algorithms via metalearning. For a distribution of related tasks, our model jointly learns: a data representation, an item selection heuristic, and a method for constructing prediction functions from labeled training sets. Our model uses the item selection heuristic to gather labeled training sets from which to construct prediction functions. Using the Omniglot and MovieLens datasets, we test our model in synthetic and practical settings.
\end{abstract}

\section{Introduction}
\label{sec:intro}

For many real-world tasks, labeled data is scarce while unlabeled data is abundant.
It is often possible, at some cost, to obtain labels for the unlabeled data.
In active learning, a model selects which instances to label so as to maximize some combination of task performance and data efficiency.
Active learning is motivated by the observation that a model may perform better while training on less labeled data if it can choose the data on which it trains~\citep{cohn1996active}.
E.g., in SVMs \citep{scholkopf2002} only the support vectors affect the decision boundary.
If one could identify the support vectors in advance, the classifier trained on the resulting set of examples would obtain the same decision boundary with less data and computation.

Active learning can benefit a variety of practical scenarios. For example, preference information for a new user in a movie recommender system may be scarce, and recommendations for the new user could be improved by carefully selecting several movies for her to rate \cite{sun2013cold,houlsby2014cold,aggarwal2016recommender}.
Likewise, collecting labels for a medical imaging task may be costly because it requires a specialist~\cite{medical}, and the cost could be reduced by carefully selecting which images to label.

Various heuristics for selecting instances to label have been proposed in the active learning literature, such as choosing the instance whose label the model is most uncertain about, or the instance whose label is expected to maximally reduce the model's uncertainty about other instances~\citep{gilad2005query,settles2010active,houlsby2011bayesian}.
We propose moving away from engineered selection heuristics towards \emph{learning} active learning algorithms end-to-end via \emph{metalearning}. Our model interacts with labeled items for many related tasks in order to learn an active learning strategy for the task at hand.
In recommendation systems, for example, ratings data for existing users can inform a strategy that efficiently elicits preferences for new users who lack prior rating data, thus \emph{bootstrapping} the system quickly out of the cold-start setting \citep{golbandi2010, golbandi2011, sun2013, kawale2015}.
A learned active learning strategy could outperform task-agnostic heuristics by sharing experience across related tasks. In particular, the model's
\begin{enumerate*}[(i)]
\item data representation, 
\item strategy for selecting items to label, and
\item prediction function constructor
\end{enumerate*}
could all co-adapt.
Moving from pipelines of independently-engineered components to end-to-end learning has lead to rapid improvements in, e.g.,~computer vision, speech recognition, and machine translation \citep{krizhevsky2012, hannun2014, he2016, wu2016}.

We base our model on the Matching Networks (MN) introduced by~\citet{vinyals2016matching}.
We extend the MN's one-shot learning ability to settings where labels are not available \emph{a priori}.
We cast active learning as a sequential decision problem: at each step the model requests the label for a particular item in a pool of unlabeled items, then adds this item to a labeled support set, which is used for MN-style prediction.
We train our model end-to-end with backprop and reinforcement learning.
We expedite the training process by allowing our model to observe and mimic a strong selection policy with oracle knowledge of the labels.

We demonstrate empirically that our proposed model learns effective active learning algorithms in an end-to-end fashion.
We evaluate the model on ``active'' variants of existing one-shot learning tasks for \emph{Omniglot}~\citep{lake2015human, vinyals2016matching, santoro2016one}, and show that it can learn efficient label querying strategies.
We also test the model's ability to learn an algorithm for bootstrapping a recommender system using the \emph{MovieLens} dataset, showing it holds promise for application in more practical settings.

\section{Related Work}
\label{sec:related}

Various heuristics have been proposed to guide the selection of which examples to label during active learning~\cite{settles2010active}. For instance, \citet{lewis1994sequential} and \citet{tong2001support} developed policies based on the confidence of the classifier, while \citet{gilad2005query} used the disagreement of a \emph{committee} of classifiers. \citet{houlsby2011bayesian} presented an approach based on Bayesian information theory,
in which examples are selected in order to maximally reduce the entropy of the posterior distribution over classifier parameters. 



The idea of learning an active learning algorithm end-to-end, via \emph{meta} active learning, was recently investigated  by~\citet{activeoneshot}. Building on the memory-augmented neural network (MANN)~\citep{santoro2016one}, the authors developed a \emph{stream-based} active learner. In stream-based active learning the model decides, while observing items presented in an exogenously-determined order, whether to predict each item's label or to pay a cost to observe its label. Our proposed model instead falls into the class of \emph{pool-based} active learners, i.e.~it has access to a static collection of unlabeled data and selects both the items for which to observe labels, and the order in which to observe them.

%
%
%
%
%
Active learning can be useful when the cost incurred for labeling an item may be traded for lower prediction error, and where the model must be data efficient (e.g.~in medical imaging~\citep{medical}). We explicitly train our model to balance between task performance and labeling cost. In this sense, we build an \emph{anytime} active learner~\cite{zilberstein1996using}, with the model trained at each step to output the best possible prediction on the evaluation set.

Our model builds on the matching-networks (MN) architecture presented by~\citet{vinyals2016matching}, which enables ``one-shot'' learning, i.e.~learning the appearance of a class from just a single example of that class~\cite{santoro2016one,koch2015siamese}.~\citet{vinyals2016matching} assume that at least one example per class exists in the labeled support set available to the model. Confronted with the harder task of composing a labeled support set from a larger pool of unlabeled examples, we show that the active learning policy learnt by our model obtains, in some cases, an equally effective support set. As in the recent one-shot learning work of \citet{santoro2016one} and \citet{vinyals2016matching}, and the active learning work of \citet{activeoneshot}, we evaluate our model on the \emph{Omniglot} dataset. This dataset was developed for the foundational one-shot learning work of \citet{lake2015human}, which focused on probabilistic program induction.

The cold-start problem is ubiquitous in recommendation systems~\citep{aggarwal2016recommender,rs1,harpale2008personalized,sun2013cold,elahi2016survey}. Instead of bootstrapping from a cold-start by randomly selecting items for a user to rate, an active learner asks for particular items to help learn a strong user model more quickly. In model-free strategies~\cite{rashid2008learning}, items are selected according to general empirical statistics such as popularity or informativeness. These approaches are computationally cheap, but lack the benefits of adaptation and personalization.
Proposals for learning an adaptive selection strategy have been made in the form of Bayesian methods that learn the parameters of a user model~\cite{houlsby2014cold,harpale2008personalized}, and in the form of decision-trees learned from existing ratings~\cite{sun2013cold}.
An extensive review can be found in~\citet{elahi2016survey}.
Intuitively, our model learns a compact, parametric representation of a decision tree end-to-end, by directly maximizing task performance. We evaluate our active learner on MovieLens-20M, a standard dataset for recommendation tasks.

We provide hints to our model during training using samples from an oracle policy that knows all the labels. Related approaches have been explored in previous work on imitation learning and learning to search \cite{ross2014, chang2015}. These methods, which focus the cost of sampling from the oracle policy on states visited by the model policy, have recently been adopted by researchers working with deep networks for representation learning \cite{zhang2017, sun2017}.

\section{Model Description}
\label{sec:model}

We now present our model, which \emph{metalearns} algorithms for active learning. Our model metalearns by attempting to actively learn on tasks sampled from a distribution over tasks, using supervised feedback to improve its expected performance on new tasks drawn from a similar distribution.
Succinctly, our model solves each task by adaptively selecting items for the labeled support set used by a Matching Network \cite{vinyals2016matching} to classify test items.
The full support set from which our model selects these examples contains both labeled and unlabeled data.

For a summary of our model, see the architecture diagram in Figure~\ref{fig:model_diagrams}, the optimization objectives in Equations~\ref{eq:task_obj_general}, \ref{eq:task_obj_training}, and \ref{eq:task_obj_gradient}, and the pseudo-code in Algorithm~\ref{alg:al_loop}. We present a formal description of our meta active learning task in Section~\ref{sec:model_task}. We describe the details of our model in Section~\ref{sec:model_architecture}, and our approach to parameter optimization in Section~\ref{sec:model_training}.

\subsection{Task Description}
\label{sec:model_task}

Our model refines its behaviour over many \emph{training episodes}, in order to maximize performance during \emph{test episodes} not encountered in training.
In each episode, our model interacts with a support set $S \equiv \{(x, y)\}$ comprising items $x$ for which the model can request labels $y$, and a similarly-defined evaluation set $E \equiv \{(\hat x, \hat y)\}$.
Let $S^u_t \equiv \{(x, \cdot)\}$ denote the set of items in the support set whose labels are still unknown after $t$ label queries, and let $S^k_t \equiv \{(x, y)\}$ denote the complementary set of items whose labels are known.
Let $S_t$ denote the joint set of labeled and unlabeled items after $t$ label queries.
Let the real-valued vector $h_t$ denote the \emph{control state} of our model after viewing $t$ labels, and let $R(E, S_t, h_t)$ denote the reward won by our model when predicting labels for the evaluation set based on the information it has received after $t$ label queries. We assume all functions depend on the model parameters $\theta$, and omit this dependence from our notation for brevity.

We define the prediction reward as follows:
\begin{equation}
R(E, S_t, h_t) \equiv \sum_{(\hat x, \hat y) \in E} \log p(\hat y | \hat x, h_t, S_t),
\label{eq:task_reward_general}
\end{equation}
which gives log-likelihood of the predictions $p(\hat y | \hat x, h_t, S_t)$ on the evaluation set. The prediction $\hat{y}$ conditions on: the test item $\hat{x}$, the current control state $h_t$, and the current labeled/unlabeled support set $S_t$. For tests on Omniglot (see Section~\ref{sec:omniglot}), we use negative cross-entropy on the class labels, and for MovieLens (see Section~\ref{sec:movielens}), we use the negative Root Mean Squared Error (RMSE).

At each step $t$ of active learning, the model requests the label for an item $x$ from the set $S^u_{t-1}$ and updates its control state from $h_{t-1}$ to $h_t$ based on the response.
Together, $h_t$ and $S_t$ determine the model's predictions for test items and the model's decision about which label to request next.
Algorithm~\ref{alg:al_loop} describes this process in detail, and Section~\ref{sec:model_architecture} formally describes the functions used in Algorithm~\ref{alg:al_loop}.

The idealized objective for training our model is:
\begin{equation}
\maximize_{\theta} \expect_{(S, E) \sim \DD} \left[ \expect_{\pi(S, T)} \left[ \sum_{t=1}^{T} R(E, S_t, h_t) \right] \right],
\label{eq:task_obj_general}
\end{equation}
in which $T$ is the max number of label queries to perform, $(S, E)$ indicates an episode sampled from some distribution $\DD$, and $\pi(S, T)$ indicates unrolling the model's active learning policy $\pi$ for $T$ steps on support set $S$. Unrolling $\pi$ produces the intermediate states $\{(S_1, h_1), ..., (S_T, h_T)\}$.

To optimize this objective, our model repeatedly samples an episode $(S, E)$, then unrolls $\pi$ for $T$ steps of active learning, and maximizes the prediction reward $R(E, S_t, h_t)$ at each step.
Alternately, our model could maximize only the reward at the final step.
We maximize reward at each step in order to promote \emph{anytime} behaviour -- i.e.~the model should perform as well as possible after each label query.
Anytime behaviour is desirable in many practical settings, e.g.~for movie recommendations the model should be robust to early termination while eliciting the preferences of a new user.

During training, for computational efficiency, our model maximizes the following approximation of Equation~\ref{eq:task_obj_general}:
\begin{align}
\expect_{(S, E) \sim \DD} \left[  \expect_{\pi(S, T)} \left[ \sum_{t=1}^{T} \tilde{R}(S^u_t, S_t, h_t) + R(E, S_T, h_T) \right] \right],
\label{eq:task_obj_training}
\end{align}
in which $\tilde{R}(S^u_t, S_t, h_t)$ is a prediction reward for unlabeled items in the support set. We assume labels for the full support set are available during training.
We compute $\tilde{R}(S^u_t, S_t, h_t)$ using a \emph{fast prediction} module, and compute $R(E, S_t, h_t)$ using a \emph{slow prediction} module.
The fast and slow prediction rewards can be obtained by substituting the appropriate predictions into Equation~\ref{eq:task_reward_general}.
Sections~\ref{sec:fast_prediction} and \ref{sec:slow_prediction} describe these modules in detail. 


\subsection{Model Architecture Details}
\label{sec:model_architecture}

\begin{figure*}[t]
\centering
\includegraphics[width=0.68\linewidth]{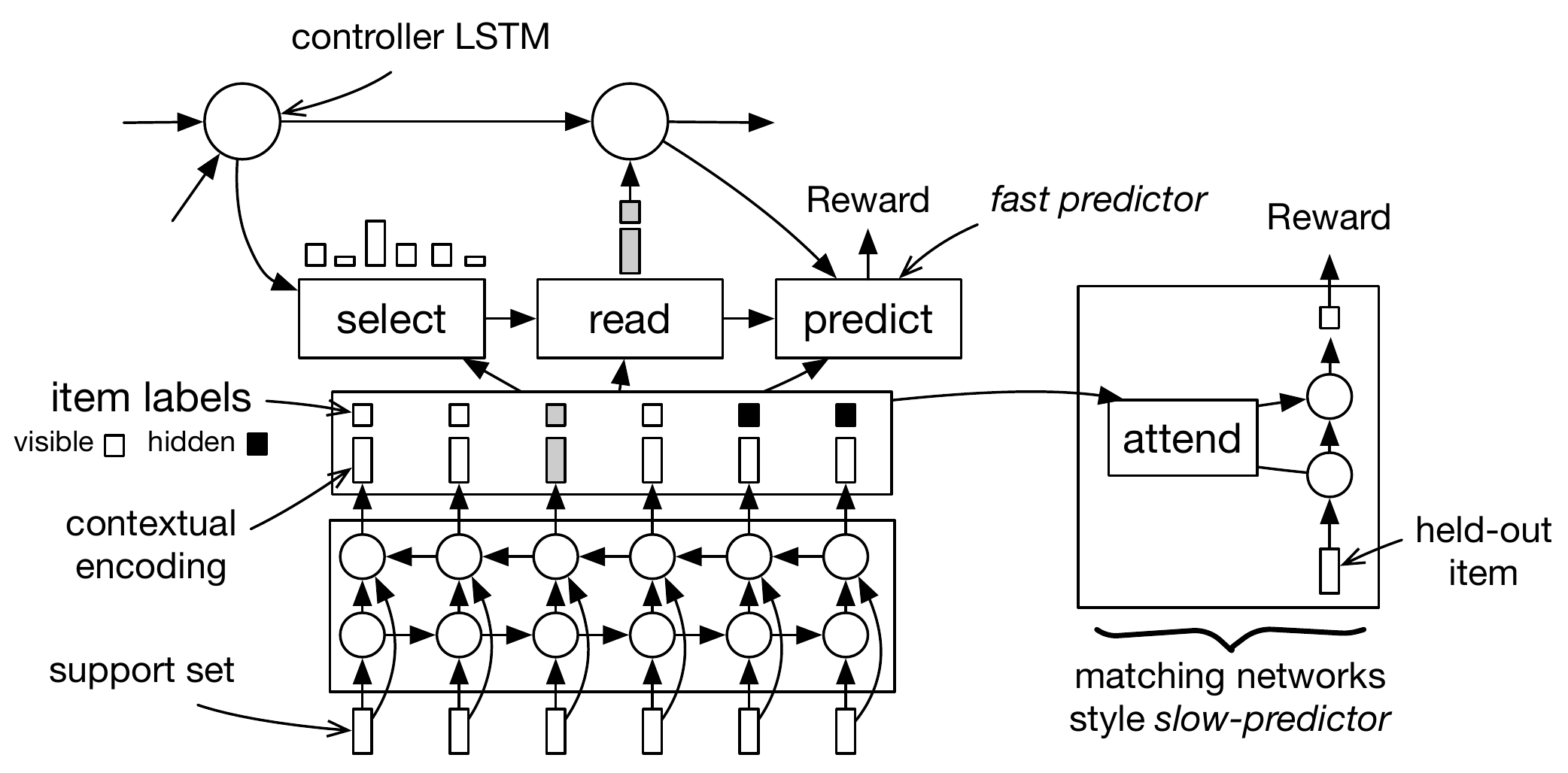}
\caption{A summary of the modules in our model. Items in the support and evaluation set are embedded using a context-free encoder. Final embeddings for support set items are computed by processing their context-free embeddings with a context-sensitive encoder. The selection module places a distribution over unlabelled items in $S^u_t$ using a gated combination of controller-item similarity features and item-item similarity features. The read module copies the selected item and its label, and transforms them for input to the controller, which then updates its state from $h_{t-1}$ to $h_{t}$. Fast predictions are made within the support set $S$ based on sharpened item-item similarity features. Slow predictions are made for items in the held-out set $E$ using a Matching Network-style function which incorporates masking to account for known/unknown labels, and conditions on the state $h_t$. We train this system end-to-end with Reinforcement Learning.}
\label{fig:model_diagrams}
\vspace{-0.5cm}
\end{figure*}

Our model comprises multiple modules: \emph{context-free} and \emph{context-sensitive} encoding, \emph{controller}, \emph{selection}, \emph{reading}, \emph{fast prediction}, and \emph{slow prediction}. We present an overview of our model in Fig.~\ref{fig:model_diagrams} and Alg.~\ref{alg:al_loop}, which describe how our model's modules perform active learning. The rest of this subsection describes the individual modules in more detail.

%
%
%

\begin{algorithm}[H]
\begin{algorithmic}[1]
\STATE \# \emph{encode items in S with context-sensitive encoder}
\STATE \# \emph{and encode items in E with context-free encoder}
\STATE $S = \{(x, y)\}$, $S_0^u = \{(x, \cdot)\}$, $S_0^k=\emptyset$, $E = \{(\hat x, \hat y)\}$
\FOR{$t = 1 \ldots T$}
\STATE \emph{\# select next instance}
\STATE $i \gets \texttt{SELECT}(S_{t-1}^u, S_{t-1}^k, h_{t-1})$ 
\STATE \emph{\# read labeled instance and update controller}
\STATE $(x_{i}, y_{i}) \gets \texttt{READ}(S, i)$
\STATE $h_{t} \gets \texttt{UPDATE}(h_{t-1}, x_i, y_i)$
\STATE \emph{\# update known / unknown set}
\STATE $S_{t}^k \gets S_{t-1}^k \cup \{(x_i, y_i)\}$
\STATE $S_{t}^u \gets S_{t-1}^u \setminus \{(x_i, \cdot)\}$
\STATE \emph{\# perform fast prediction (save loss for training)}
\STATE $L_{t}^S \gets \texttt{FAST-PRED}(S, S_{t}^u, S_{t}^k, h_{t})$
\ENDFOR
\STATE \emph{\# perform slow prediction (save loss for training)}
\STATE $L_T^E \gets \texttt{SLOW-PRED}(E, S_{T}^u, S_{T}^k, h_T)$
\end{algorithmic}
\caption{End-to-end active learning loop (for Eq.~\ref{eq:task_obj_training})}
\label{alg:al_loop}
\end{algorithm}

\subsubsection{Context-[Free|Sensitive] Encoding}
The context-free encoder associates each item with an embedding independent of the context in which the item was presented. For our Omniglot tests, this encoder is a convnet with two convolutional layers that downsample via strided convolution, followed by another convolutional layer and a fully-connected linear layer which produces the final context-free embedding. For our MovieLens tests, this encoder is a simple look-up-table mapping movie ids to embeddings. We denote the context-free encoding of item $x_i \in S$ as $x^{\prime}_i$, and similarly define $\hat{x}^{\prime}_i$ for $\hat{x}_i \in E$.

The context-sensitive encoder produces an embedding $x_i^{\prime\prime}$ for each item $x_i \in S$ based on the context-free embeddings $x_j^{\prime}: \; \forall x_j \in S$. The context-sensitive encoder is not applied to items in the evaluation set. Our model uses a modified form of the encoder from Matching Networks \cite{vinyals2016matching}. Specifically, we run a bidirectional LSTM \cite{hochreiter1997, Schuster1997} over all context-free embeddings for the support set, and then add a linear function of the concatenated forwards and backwards states to the context-free embedding $x_i^{\prime}$ to get $x_i^{\prime\prime}$.

We can write this as follows:
\begin{equation}
x_i^{\prime\prime} = x_i^{\prime} + W_{e} \left[ \vec{h}_i; \cev{h}_i \right],
\end{equation}
in which $\vec{h}_i$ gives the forward encoder state for item $x_i$, $\cev{h}_i$ gives the backward encoder state, and $W_{e}$ is a trainable matrix. We compute the forward states $\vec{h}_i$ as in a standard LSTM, processing the support set items $x_i$ sequentially following a random order. We compute the backward states $\cev{h}_i$ by processing the sequence of concatenated $x_i^{\prime}$ and $\vec{h}_i$ vectors in reverse.

\subsubsection{Reading}
This module concatenates the embedding $x^{\prime\prime}_i$ and label $y_i$ for the item indicated by the selection module, and linearly transforms them before passing them to the controller (Alg.~\ref{alg:al_loop}, line 8).

\subsubsection{Controller}
At each step $t$, the controller receives an input $r_t$ from the reading module which encodes the most recently read item/label pair. Additional inputs could take advantage of task-specific information. The control module performs an LSTM update:
$$h_t = \mbox{LSTM}(h_{t-1}, r_t).$$
We initialize $h_0$ for each episode $(S, E)$ using the final state of the backwards LSTM in the context-sensitive encoder which processed the support set $S$. In principal, this allows the controller to condition its behaviour on the full unlabeled contents of the support set  (Alg.~\ref{alg:al_loop}, line 9).

\subsubsection{Selection}
At each step $t$, the selection module places a distribution $P^u_t$ over all unlabeled items $x^u_i \in S^u_t$. It then samples the index of an item to label from $P^u_t$, and feeds it to the reading module (Alg.~\ref{alg:al_loop}, line 6).

Our model computes $P^u_t$ using a gated, linear combination of features which measure controller-item similarity and item-item similarity. For each item, we compute the controller-item similarity features:
$$b^i_t = x^{\prime\prime}_i \odot W_{b} h_t,$$
where $W_{b}$ is a trainable matrix and $\odot$ indicates elementwise multiplication.
We also compute the following six item-item similarity features: [max|mean|min] cosine similarity to any labeled item, and [max|mean|min] cosine similarity to any unlabeled item.
We concatenate the controller-item similarity features and item-item similarity features to get a vector $d^i_t$. We also compute a gating vector: $g_t = \sigma(W_g h_t)$, in which $W_g$ is a trainable matrix and $\sigma(\cdot)$ indicates the standard sigmoid.

For each $x^u_i \in S^u_t$, we compute the selection logit: $$p^i_t = (g_t \odot d^i_t)^{\top} w_p,$$ where $w_p$ indicates a trainable vector. Finally, we compute $P^u_t$ by normalizing over the logits $p^i_t: \; \forall x^u_i \in S^u_t$ via softmax. This module performs worse when the controller-item or item-item features are removed. Our model intelligently adapts these heuristics to the task at hand. We provide pseudo-code for how these modules interact during active learning in Algorithm~\ref{alg:al_loop}.

\subsubsection{Fast Prediction}
\label{sec:fast_prediction}
The fast prediction module makes an attention-based prediction for each unlabeled item $x^u_i \in S^u_t$ using its cosine similarities to the labeled items $x^k_j \in S^k_t$, which are sharpened by a non-negative matching score $\gamma^i_t$ between $x^u_i$ and the control state $h_t$. The cosine similarities are taken between the context-sensitive embeddings $x^{\prime\prime}_i$ and $x^{\prime\prime}_j$ of the respective items. These do not change with $t$ and may be precomputed before unrolling the active learning policy. Predictions from this module are thus fast to compute while unrolling the policy (Alg.~\ref{alg:al_loop}, line 14). The cosine similarities may be reused in the selection module for computing item-item similarity features, further amortizing their cost.

For each unlabeled $x^u_i$, we compute a set of attention weights over the labeled $x^k_j \in S^k_t$ by applying a softmax to the relevant cosine similarities, using $\gamma^i_t$ as a temperature for the softmax. We compute the sharpening term as follows:
$$\gamma^i_t = \exp((x^{\prime\prime}_i)^{\top} W_{\gamma} h_t),$$
where $W_{\gamma}$ indicates a trainable matrix.
This module performs significantly worse without the sharpening term.
The final \emph{fast prediction} is formed by taking a convex combination of the labels $y_j$ for the labeled $x^k_j \in S^k_t$ using the computed attention weights.

\subsubsection{Slow Prediction}
\label{sec:slow_prediction}
The slow prediction module implements a modified Matching Network prediction, which accounts for the distinction between labeled and unlabeled items in $S_t$, and conditions on the active learning control state $h_t$ (Alg.~\ref{alg:al_loop}, line 17).

Given the context-free embedding $\hat{x}^{\prime}$ for some held-out example $\hat{x} \in E$, the state $h_t$, and required initial values,
this module predicts a label by iterating the steps:
\begin{enumerate}
\setlength\itemsep{0.01em}
\item $m_k = \mbox{LSTM}(m_{k-1}, \tilde{x}_{k-1}, \hat{x}^{\prime}, h_t)$
\item $\hat{x}^{\prime\prime} = \hat{x}^{\prime} + W_m m_k$
\item $\tilde{a}_k = \mbox{attend}(\hat{x}^{\prime\prime}, S^k_t)$
\item $\tilde{x}_k, \tilde{y}_k = \mbox{attRead}(\tilde{a}_k, S^k_t)$
\end{enumerate}
Here, $\mbox{LSTM}$ is an LSTM update, $m_k$ is the \emph{matching state} at step $k$, $\hat{x}^{\prime\prime}$ is the match-sensitive embedding of $\hat{x}$ at step $k$, $W_m$ is a trainable matrix, $\tilde{a}_k$ are the matching attention weights at step $k$, $\tilde{x}_k$ is the ``item attention'' result from step $k$, and $\tilde{y}_k$ is the ``label attention'' result from step $k$.
For details of the $\mbox{attend}$ and $\mbox{attRead}$ functions, refer to~\citet{vinyals2016matching}.
As a final prediction, this module returns the label attention result $\tilde{y}_K$ from the $K$th (final) step of iterative matching. In our tests we fix $K = 3$.

\textbf{Note:} our model contains many linear transforms $W v$. Our model adds bias and gain terms to all of these transforms, as described for weight normalization \cite{Salimans2016}. We omit these terms for brevity. Similarly, we use layer normalization in our active learning and matching controller LSTMs \cite{Ba2016}.

\subsection{Training the Model}
\label{sec:model_training}

We optimize the parameters of our model using a combination of backpropagation and policy gradients. For a clear review of optimization techniques for general stochastic computation graphs, see \citet{schulman2015}.

Using the notation from Section~\ref{sec:model_task} and following the approach of \cite{schulman2015}, we can write the gradient of our training objective as follows:
\begin{align}
 \nabla_{\theta} R(& S, E, \theta) = \label{eq:task_obj_gradient} \\
 \expect_{p(\vec{S} | (S, E))} \bigg(& \nabla_{\theta} \log p(\vec{S} | (S, E)) \left[ R(\vec{S}) \right] + \nabla_{\theta} R(\vec{S}) \bigg), \nonumber
\end{align}
in which $R(S, E, \theta)$ denotes the expected reward won by the active learning model while working on episode $(S, E)$. $\vec{S}$ denotes the set of intermediate states $\{(S_t, h_t)\}$ generated by the model while working with $(S, E)$. $R(\vec{S})$ denotes the sum of rewards (as described in Equation~\ref{eq:task_obj_training}) received by the model while working on episode $(S, E)$. In the term $\nabla_{\theta} R(\vec{S})$, all decisions made by the model to produce $\vec{S}$ are treated as constant. Taking the expectation of Equation~\ref{eq:task_obj_gradient} with respect to episodes $(S, E) \sim \DD$ gives us an unbiased gradient estimator for the objective in Equation~\ref{eq:task_obj_training}.

Rather than using the gradients in Equation~\ref{eq:task_obj_gradient} directly, we optimize the model parameters using Generalized Advantage Estimation \cite{schulman2016}, which provides an actor-critic approach for approximately optimizing the policy gradients in Equation~\ref{eq:task_obj_gradient}.
For more details on how Generalized Advantage Estimation helps reach a favourable bias-variance trade-off in policy gradient estimation, see the source paper \cite{schulman2016}. We apply the GAE updates using ADAM \cite{kingma2015}.

\section{Experiments}
\label{sec:exp}
\subsection{Omniglot}
\label{sec:omniglot}


\begin{table*}[]
\centering
\caption{Results for our active learner and baselines for the $N$-way, $K$-shot classification settings.}
\label{tab:res_kway_kshot}
\begin{tabular}{@{}lllllll@{}}
\toprule
\multicolumn{1}{c}{\multirow{2}{*}{Model}} & \multicolumn{3}{c}{\textbf{5-way}} & \multicolumn{3}{c}{\textbf{10-way}} \\ \cmidrule(l){2-7} 
\multicolumn{1}{c}{}                       & 1-shot     & 2-shot    & 3-shot    & 1-shot     & 2-shot     & 3-shot    \\ \midrule
\textbf{Matching Net (random)} & 69.8\%$_{\pm 0.10}$     & 93.1\%$_{\pm 0.07}$    & 98.5\%$_{\pm 0.04}$ & 67.3\%$_{\pm 0.10}$  & 91.2\%$_{\pm 0.06}$ & 97.6\%$_{\pm 0.06}$    \\
\textbf{Matching Net (balanced)} & 97.9\%$_{\pm 0.07}$     & 98.9\%$_{\pm 0.07}$    & 99.2\%$_{\pm 0.06}$    & 96.5\%$_{\pm 0.04}$     & 98.3\%$_{\pm 0.03}$    & 98.7\%$_{\pm 0.05}$    \\
\textbf{Active MN} & 97.4\%$_{\pm 0.11}$     & 99.0\%$_{\pm 0.08}$    & 99.3\%$_{\pm 0.03}$    & 94.3\%$_{\pm 0.24}$    & 98.0\%$_{\pm 0.07}$     & 98.5\%$_{\pm 0.06}$    \\
\midrule
\textbf{Min-Max-Cos} & 97.4\%$_{\pm 0.11}$     & 99.3\%$_{\pm 0.02}$    & 99.4\%$_{\pm 0.04}$    & 93.5\%$_{\pm 0.11}$    & 98.4\%$_{\pm 0.02}$     & 98.8\%$_{\pm 0.03}$    \\ \bottomrule
\end{tabular}
\vspace{-0.25cm}
\end{table*}

\begin{figure}[t]
\begin{center}
\vspace{-0.1cm}
\hspace{-0.5cm}
\includegraphics[width=0.9\linewidth]{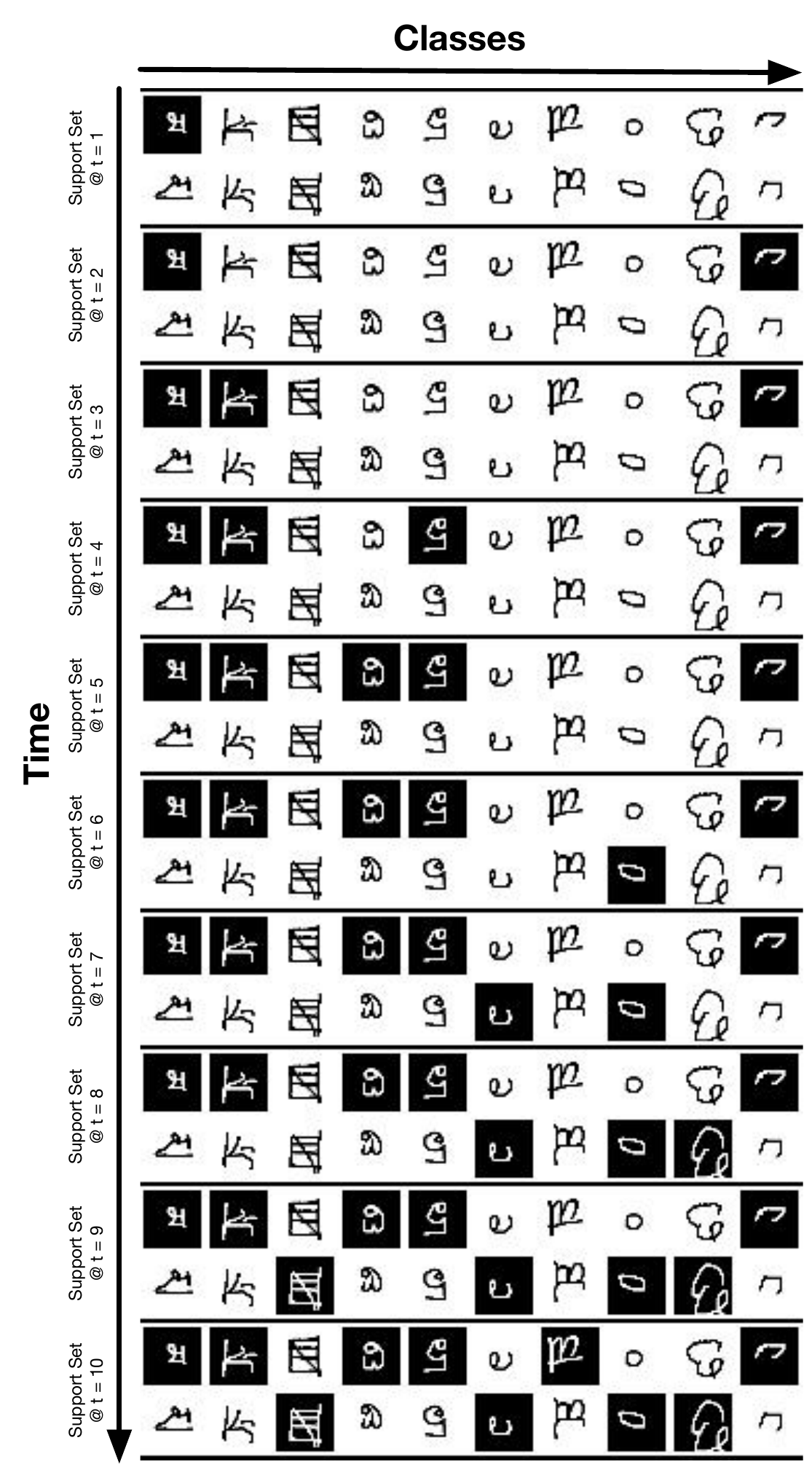}
\vspace{-0.25cm}
\caption{A rollout of our active learning policy for Omniglot, using a support set of 20 items from 10 different classes with 2 items per class. Each row represents the support set at different active iterations. For visualization purposes, each column represents a class. Unlabeled items have white background while selected items have black background. Here, the model behaves intelligently, by selecting at each step an item with a yet-unseen label.}
\label{fig:og_rollout}
\end{center}
\vspace{-0.5cm}
\end{figure}

We run our first experiments on the Omniglot dataset~\citep{lake2015human} consisting of 1623 characters from 50 different alphabets, each hand-written by 20 different people.
Following~\citet{vinyals2016matching}, we divide the dataset into 1200 characters for training and keep the rest for testing. When measuring test performance, our model interacts with characters it did not encounter during training.
%
%

For the context-free embedding function we use a three-layer convolutional network. The first two layers use $5 \times 5$ convolutions with 64 filters and downsample with a double stride. The third layer uses a $3 \times 3$ convolution with 64 filters and no downsampling. These layers produce a $7 \times 7 \times 64$ feature map that we flatten and pass through a fully connected layer. All convolutional layers use the leaky ReLU nonlinearity \cite{Maas2013}.

We setup $N$-way, $K$-shot Omniglot classification as follows. We randomly pick $N$ character classes from the available train/test classes. Then, we build a support set by randomly sampling 5 items for each character class,~e.g. in the 5-way setting, there are 25 items in the support set. The held-out set is always obtained by randomly sampling 1 item per class.
In our active learning setting, $K$-shot is proportional to how many labels the model can acquire. In the $N$-way, $K$-shot setting, the model asks for $NK$ labels before performing held-out prediction. For example, in 5-way, 1-shot classification, the model asks for 5 labels. Following each label query, we also measure anytime performance of the fast prediction module on the items remaining in $S^u_t$. Note that the 1-shot setting is particularly challenging for our model, as it needs to ask for different classes at each step, and the ability to identify missing classes is limited by the accuracy of the underlying one-shot classifier.

We compare our active learner to four baselines. To compute a pessimistic estimate of potential performance, we use a matching network where we label $NK$ items chosen at random from the full support set (Matching Net (random)). As the labels are randomly sampled, it is possible that a given class is never represented among the labeled items and the model cannot classify perfectly, even in principle. To compute an optimistic estimate of potential performance, we measure the ``ideal'' matching network accuracy by labeling a class-balanced subset of items from the full support set (Matching Net (balanced)). This baseline represents a highly-performant policy that the active learner can, in principle, learn. For the last baseline (Min-Max-Cos), we formulate a heuristic policy. At each active learning step, we select the item which has minimum maximum cosine similarity to unlabeled items in the support set. This heuristic selects item that are different from each other, a strategy well-suited to the Omniglot classification task where items are drawn from a consistent set of underlying classes.

We report the results in Table~\ref{tab:res_kway_kshot}. Matching Networks operating on a randomly sampled set of labels suffer the most in 1-shot scenarios, where the probability of all classes being represented is particularly low (especially in the 10-way case). Overall, the active policy nearly matches the performance of the optimistic balanced Matching Network baseline. Degradation of performance by 2.2\% is observed for the 1-shot, 10-way case. This is not surprising since augmenting the number of classes in the support set, while keeping the number of shots fixed, considerably increases the difficulty of the problem for the active learner.
Figure~\ref{fig:og_rollout} shows a roll-out of the model policy in the 10-way setting.

\begin{figure*}
\begin{center}
\includegraphics[width=0.38\linewidth]{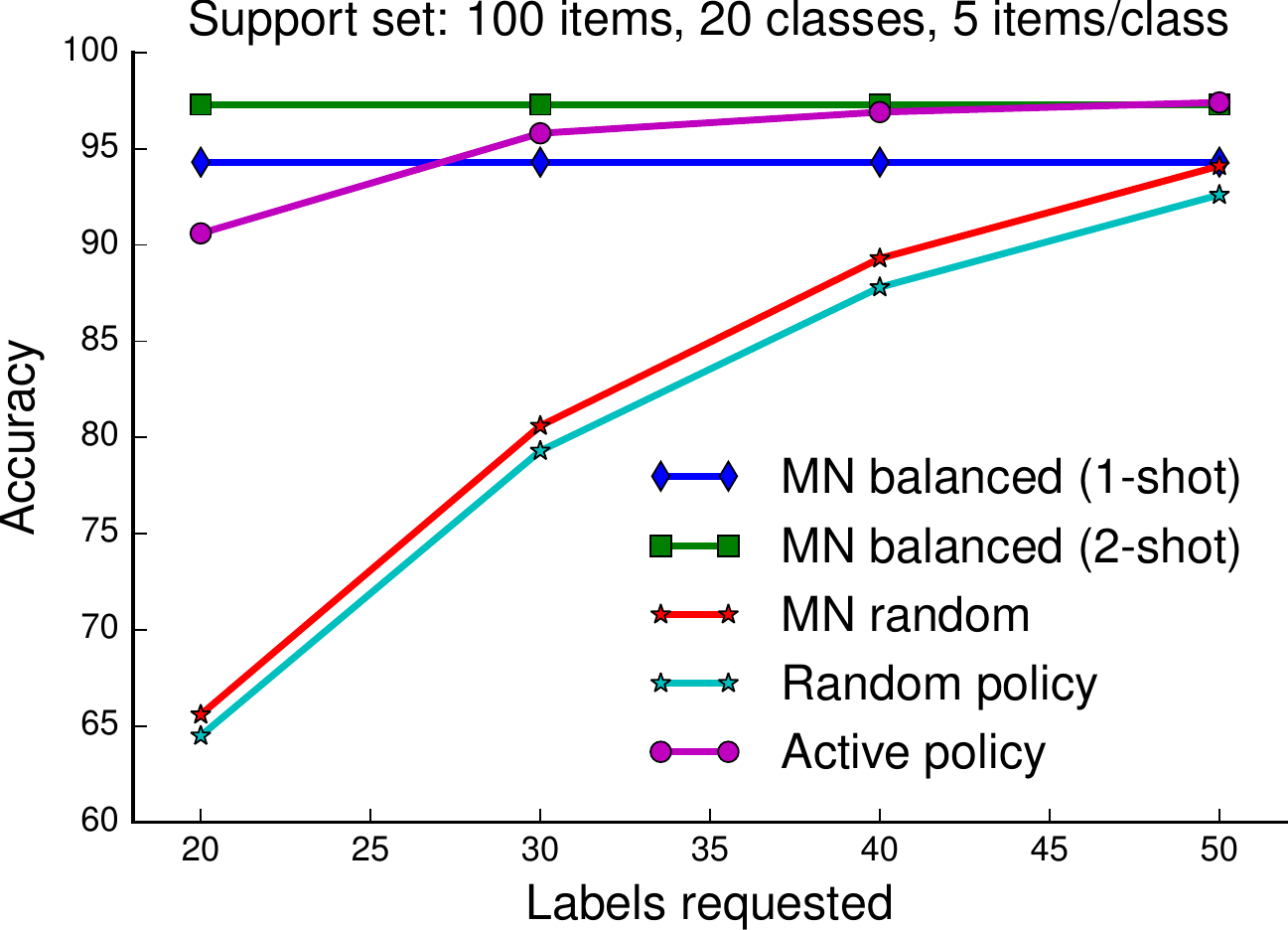}
\includegraphics[width=0.38\linewidth]{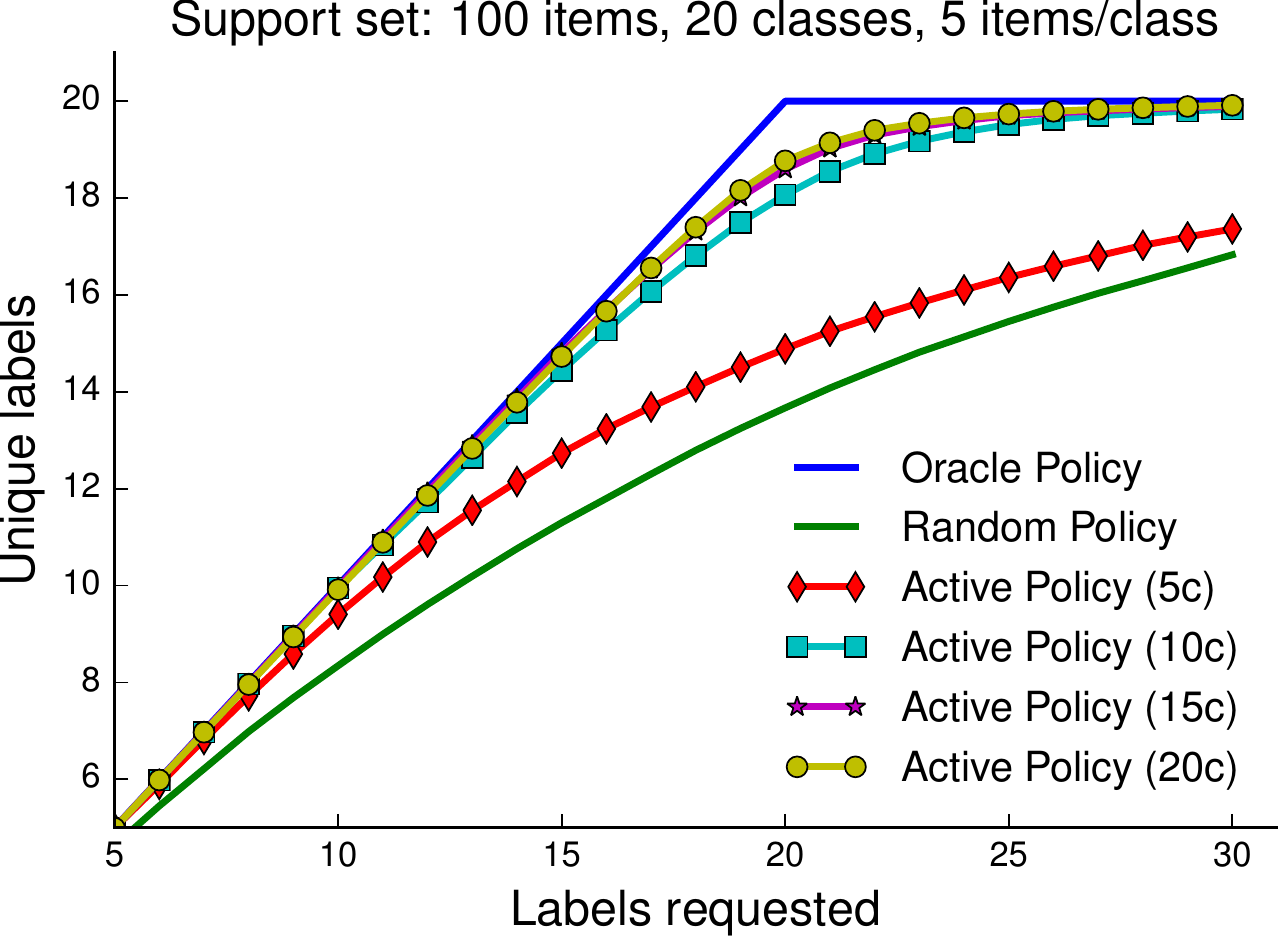}
\caption{Experiment results for our model and baselines on Omniglot. The left plot shows how prediction accuracy improves with the number of labels requested in a challenging 20-way setting. After 20 label requests~(corresponding to a 20-way, 1-shot problem), the active policy outperforms random policy and random MN baselines, but is inferior to the balanced MN. After 30 labels, the active policy nearly matches the performance of the balanced MN using 40 labels (20-way, 2-shot). The right plot shows the number of unique labels with respect to the number of requested labels for models trained on problems with 5-20 classes, and tested on 20-way classification. This gives an idea of how models search for labels from unseen groups and generalize to problems with different numbers of classes.}
\label{fig:mega_plot}
\end{center}
\vspace{-0.5cm}
\end{figure*}

Figure~\ref{fig:mega_plot} provides results for the more challenging setting of 20-way classification. We tested two properties of our model: its anytime performance beyond the 1-shot setting, and its ability to generalize to problems with more classes than were seen during training. The model performed well on 20-way classification, and quickly approached the optimistic performance estimate after acquiring more labels. We also found that policies trained for as little as 10-way classification could generalize to the 20-way setting.

Our model relies on a number of moving parts. When designing the architecture, we followed the simple approach of minimizing changes to the original Matching Network from \citet{vinyals2016matching}. We now provide ablation test results for several parts of our model. In the 10-way, 1-shot setting accuracy dropped from 94.5 to 86.0 when we removed attention temperature from the fast prediction module. Reducing the number of matching steps from 3 to 2 or 1 had no significant effect in this setting. Removing the context-sensitive encoder also had no significant effect. Streamlining our architecture is clearly a useful topic for future work aimed at scaling our approach to more realistic settings.

\subsection{MovieLens}
\label{sec:movielens}

\begin{figure}[t]
\begin{center}
\includegraphics[width=0.8\linewidth]{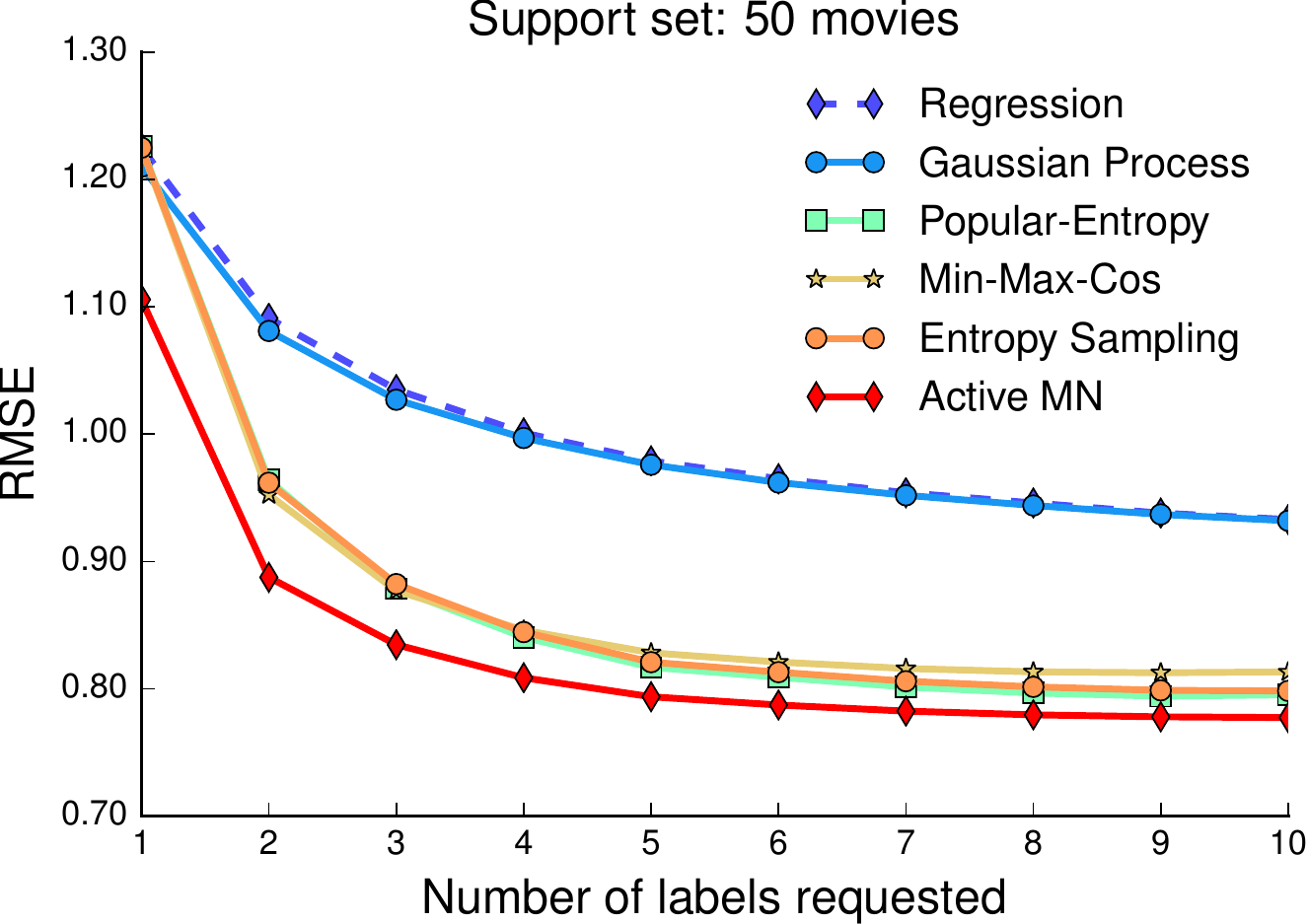}
\caption{Performance of the model and baselines measured with RMSE on the Movielens dataset.}
\label{fig:mlens_plot}
\end{center}
\vspace{-0.5cm}
\end{figure}

\subsubsection{Setup}
We test our model in the ``cold-start'' collaborative filtering scenario using the publicly available MovieLens-20M dataset.\footnote{Available at \url{http://grouplens.org/datasets/movielens/}}
The dataset contains approximately 20M ratings on 27K movies by 138K users. The ratings are on an ordinal 10-point scale, from 0.5 to 5 with intervals of 0.5. We subsample the dataset by selecting 4000 movies and 6000 users with the most ratings. After filtering, the dataset contains approximately 1M ratings. We partition the data randomly into 5000 training users and 1000 test users. The training set represents the users already in the system who are used to fit the model parameters. We use the test users to evaluate our active learning approach. For each user, we randomly pick 50 ratings to include in the support set (movies that the user can be queried about) and 10 movies and ratings for the held-out set. We ensure that movies in the held-out set and in the support set do not overlap. We train our active learner to minimize the mean-squared error (MSE) with respect to the true rating. We adapt the prediction modules of our model to output the rating for a held-out item as follows: we compute a convex combination of the ratings of ``visible'' movies in the support set (the movies the active learner has already queried about), where the weights are given by the final attention step of the slow predictor. Although more complex strategies are possible, we empirically found this simple strategy to work well in our experiments. 
For evaluation, we sample 25000 episodes~(each comprising 50 support ratings and 10 held-out ratings from some user) from the test set and we compute the average per-user root mean-squared error (RMSE). We report the average performance obtained by 3 runs with different random seeds.

\subsubsection{Movie Embeddings}
For each movie, we pretrain an embedding vector by decomposing the full user/movie rating matrix using a latent factor model~\citep{koren2010factor}. This process only uses the training set. For each user $u$ and movie $m$, we estimate the true rating $r_{u, m}$ with a linear model $\hat{r}_{u, m} = x_u^\top x_m + b_u + b_m + \beta$, where $x_u, x_m$ are the user and movie embedding respectively and $b_u, b_m, \beta$ are the user, movie, and global bias, respectively. We train the latent factor model by minimizing the mean squared error between true rating $r$ and predicted rating $\hat{r}$. We use the trained $x_m$ as input representations for the movies throughout our experiments.

\subsubsection{Results}
In Figure~\ref{fig:mlens_plot} we report the results of our active model against various baselines.
The Regression baseline performs regularized linear regression on movies from the support set whose ratings have been observed incrementally in random order.
Because of the small amount of training data, for each additional label we tune the regularization parameter by monitoring performance on a separate set of validation episodes.
The Gaussian Process baseline selects the next movie to label in proportion to the variance of the predictive posterior distribution over its rating.
This gives an idea of the impact of using MN one-shot capabilities rather than standard regression techniques.
The Popular-Entropy, Min-Max-Cos, and Entropy Sampling baselines train our model end-to-end, but using fixed selection policies.
Specifically, we train our architecture end-to-end, but instead of training an active learning policy through the select module we choose items from the support set incrementally according to a heuristic policy.
This gives an idea of the importance of learning the selection policy.
The Popular-Entropy policy, adapted from the cold-start work of~\citet{rashid2002getting}, scores each item in the support set a priori, according to the logarithm of its popularity multiplied by the entropy of the item's ratings measured across users.
This strategy aims to first collect the ratings for those movies that are both popular and have been rated differently by different users.
Although it is simplistic, the policy achieves competitive performance for bootstrapping a system from a cold-start setting~\cite{elahi2016survey}.
The Min-Max-Cos policy is identical to the synonymous baseline used for Omniglot,~i.e. it selects the unrated movie which has minimum maximum cosine similarity to any of the rated movies.
Roughly, this selects the unrated movie which differs most from the rated movies. Entropy Sampling selects movies in proportion to rating prediction entropy.

The active policy learned end-to-end outperforms the baselines in terms of RMSE, particularly after requesting only the first few labels.
After 10 ratings, our model achieves an improvement of 2.5\% in RMSE against the best baseline.
Unsurprisingly, the gap diminishes with a higher number of labels requested.
After observing 5 labels, the Popular-Entropy baseline and our architecture equipped with the Min-Max-Cos heuristic converge toward the active policy but never quite match it.
For MovieLens, where labels are user-dependent and not tied to an underlying class, a data-driven selection policy may adapt better to the task.
This contrasts with the Omniglot setting, where there is no aspect of personalization and Active MN and Min-Max-Cos perform similarly.
The Min-Max-Cos heuristic is designed to not select items similar to those it has already seen, but selecting similar items can be beneficial in personalized settings~\cite{elahi2016survey}.


\section{Conclusion}
\label{sec:conc}

We introduced a model that learns active learning algorithms end-to-end. Our goal was to move away from engineered selection heuristics towards strategies learned directly from data. Our model leverages labeled instances from different but related tasks to learn a selection strategy for the task at hand, while simultaneously adapting its representation of the data and its prediction function.
We evaluated the model on ``active'' variants of one-shot learning tasks for Omniglot, demonstrating that its policy approaches an optimistic performance estimate. On a cold-start collaborative filtering task derived from MovieLens, the model outperforms several baselines and shows promise for application in more realistic settings.

{\footnotesize
\bibliography{ggrefs}
\bibliographystyle{icml2017}
}

\end{document}